\documentclass[11pt,a4paper]{article}
\usepackage[hyperref]{acl2017}
\usepackage{times}
\usepackage{latexsym}
\usepackage{graphicx}
\usepackage{lipsum}
\usepackage{multirow}
\usepackage{bm}
\usepackage{amsmath}

\usepackage{url}

\aclfinalcopy 

\title{Sequential Dialogue Context Modeling for Spoken Language Understanding}

\author{Ankur Bapna \\\\
  {\tt \quad \quad \quad \quad ankurbpn@google.com} \\\And
  Gokhan T{\"u}r \\
  \quad \quad \quad \quad \quad \quad \quad \quad Google Research, Mountain View\\\And
  Dilek Hakkani-T{\"u}r \\\\
  {\tt \{gokhan.tur, dilek, larry.heck\}@ieee.org} \\\And
  Larry Heck}

\date{}

\begin{document}
\maketitle
\begin{abstract}
Spoken Language Understanding (SLU) is a key component of goal oriented dialogue systems that would parse user utterances into semantic frame representations. Traditionally SLU does not utilize the dialogue history beyond the previous system turn and contextual ambiguities are resolved by the downstream components. In this paper, we explore novel approaches for modeling dialogue context in a recurrent neural network (RNN) based language understanding system. We propose the Sequential Dialogue Encoder Network, that allows encoding context from the dialogue history in chronological order. We compare the performance of our proposed architecture with two context models, one that uses just the previous turn context and another that encodes dialogue context in a memory network, but loses the order of utterances in the dialogue history. Experiments with a multi-domain dialogue dataset demonstrate that the proposed architecture results in reduced semantic frame error rates.
\end{abstract}

\section{Introduction}
Goal oriented dialogue systems help users with accomplishing tasks, like making restaurant reservations or booking flights, by interacting with them in natural language. The capability to understand user utterances and break them down into task specific semantics is a key requirement for these systems. This is accomplished in the spoken language understanding module, which typically parses user utterances into semantic frames, composed of domains, intents and slots~\cite{tur2011spoken}, that can then be processed by downstream dialogue system components. An example semantic frame is shown for a restaurant reservation related query in Figure~\ref{fig:ex}. \\
As the complexity of the task supported by a dialogue system increases, there is a need for an increased back and forth interaction between the user and the agent. For example, a restaurant reservation task might require the user to specify a restaurant name, date, time and number of people required for the reservation. Additionally, based on reservation availability, the user might need to negotiate on date, time, or any other attribute with the agent. This puts the burden of parsing in-dialogue contextual user utterances on the language understanding module. The complexity increases further when the system supports more than one task and the user is allowed to have goals spanning multiple domains within the same dialogue. Natural language utterances are often ambiguous, and the context from previous user and system turns could help resolve the errors arising from these ambiguities.
\begin{figure}[t]
  \centering
  \begin{tabular}{|p{0.25cm}p{6.3cm}|}
    \hline
  	$u1$&Can you get me a restaurant reservation ? \\
  	$s$&Sure, where do you want to go ?\\
  \end{tabular}
  \begin{tabular}{|ccccccc|}
    $u2$&table&for&2&at&Pho&Nam\\
    &$\downarrow$&$\downarrow$&$\downarrow$&$\downarrow$&$\downarrow$&$\downarrow$\\
    $S$&O&O&B-\#&O&B-Rest&I-Rest\\
  \end{tabular}
  \begin{tabular}{|p{0.55cm}p{6cm}|}
    $D$&restaurants\\
    $I$&reserve\_restaurant\\
    \hline
  \end{tabular}
  \caption[Figure]{An example semantic parse of an utterance ($u2$) with slot ($S$), domain ($D$), intent ($I$) annotations, following the IOB (in-out-begin) representation for slot values.}
  \label{fig:ex}
\vspace{-0.5cm}
\end{figure}
\begin{figure*}
  \centering
	\includegraphics[width=0.6\textwidth]{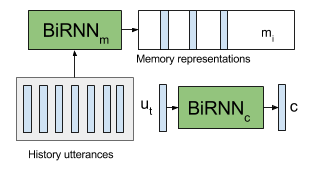}
	\caption{Architecture of the Memory and current utterance context encoder.}
	\label{fig:memcontext}
\vspace{-0.5cm}
\end{figure*}

In this paper, we explore approaches to improve dialogue context modeling within a Recurrent Neural Network (RNN) based spoken language understanding system. 
We propose a novel model architecture to improve dialogue context modeling for spoken language understanding on a multi-domain dialogue dataset. The proposed architecture is an extension of Hierarchical Recurrent Encoder Decoders (HRED)~\cite{HRED2015}, where we combine the query level encodings with a representation of the current utterance, before feeding it into the session level encoder. We compare the performance of this model to a RNN tagger injected with just the previous turn context and a single hop memory network that uses an attention weighted combination of the dialogue context~\cite{vivianIS16,weston2014memory}. \\
Furthermore, we describe a dialogue recombination technique to enhance the complexity of the training dataset by injecting synthetic domain switches, to create a better match with the mixed domain dialogues in the test dataset. This is, in principle, a multi-turn extension of ~\cite{jia2016data}. Instead of inducing and composing grammars to synthetically enhance single turn text, we combine single domain dialogue sessions into multi-domain dialogues to provide richer context during training.

\section{Related Work}
The task of understanding a user utterance is typically broken down into 3 tasks: domain classification, intent classification and slot-filling~\cite{tur2011spoken}.
Most modern approaches to Spoken language understanding involve training machine learning models on labeled training data~\cite[among others]{young,luna,WangDengAcero05}. More recently, recurrent neural network (RNN) based approaches have been shown to perform exceedingly well on spoken language understanding tasks~\cite[among others]{RNN-TASL,dilekIS16,Kurata:emnlp16}. RNN based approaches have also been applied successfully to other tasks for dialogue systems, like dialogue state tracking ~\cite[among others]{matthen,henderson2014second,perez2016dialog}, policy learning ~\cite{su2015reward} and system response generation ~\cite[among others]{wen2015semantically,wen2016multi}. \\
In parallel, joint modeling of tasks and addition of contextual signals has been shown to result in performance gains for several applications. Modeling domain, intent and slots in a joint RNN model was shown to result in reduction of overall frame error rates~\cite{dilekIS16}. Joint modeling of intent classification and language modeling showed promising improvements in intent recognition, especially in the presence of noisy speech recognition~\cite{LiuL16e}.\\
Similarly, models incorporating more context from dialogue history~\cite{vivianIS16} or semantic context from the frame~\cite{Yann-ICLR,ankur:IS17} tend to outperform models without context and have shown potential for greater generalization on spoken language understanding and related tasks. ~\cite{dhingra2016end} show improved performance on an informational dialogue agent by incorporating knowledge base context into their dialogue system. Using dialogue context was shown to boost performance for end to end dialogue ~\cite{bordes2016learning} and next utterance prediction ~\cite{SerbanSBCP15}. \\
In the next few sections, we describe the proposed model architecture, the dataset and our dialogue recombination approach. This is followed by experimental results and analysis.

\section{Model Architecture}
\begin{figure*}
  \centering
	\includegraphics[width=0.75\textwidth]{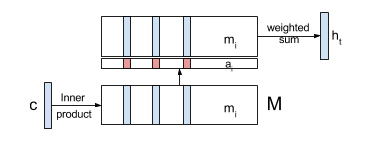}
	\caption{Architecture of the dialogue context encoder for the cosine similarity based memory network.}
	\label{fig:cosinemem}
\vspace{-0.5cm}
\end{figure*}
We compare the performance of 3 model architectures for encoding dialogue context on a multi-domain dialogue dataset. Let the dialogue be a sequence of system and user utterances $D_t = \{u_1, u_2 ... u_t\}$ and at time step $t$ we are trying to output the parse of a user utterance $u_t$, given $D_t$. Let any utterance $u_k$ be a sequence of tokens given by $\{x^k_1, x^k_2 ... x^k_{n^k}\}$. \\
We divide the model into 2 components, the context encoder that acts on $D_t$ to produce a vector representation of the dialogue context denoted by $h_t = H(D_t)$, and the tagger, which takes the dialogue context encoding $h_t$, and the current utterance $u_t$ as input and produces the domain, intent and slot annotations as output.
\subsection{Context Encoder Architectures}
In this section we describe the architectures of the context encoders used for our experiments. We compare the performance of 3 different architectures that encode varying levels of dialogue context.
\subsubsection{Previous Utterance Encoder}
This is the baseline context encoder architecture. We feed the embeddings corresponding to tokens in the previous system utterance, $u_{t-1} = \{x^{t-1}_1, x^{t-1}_2 ... x^{t-1}_{n^{t-1}}\}$, into a single Bidirectional RNN (BiRNN) layer with Gated Recurrent Unit (GRU) ~\cite{chung2014empirical} cells and 128 dimensions (64 in each direction). The embeddings are shared with the tagger. The final state of the context encoder GRU is used as the dialogue context. 
\begin{equation} \label{eq:luecontext}
h_t = BiGRU_c(\bm{u_{t-1}})
\end{equation}
\subsubsection{Memory Network}
This architecture is identical to the approach described in ~\cite{vivianIS16}. We encode all dialogue context utterances, $\{u_1, u_2 ... u_{t-1}\}$, into memory vectors denoted by $\{m_1, m_2, ... m_{t-1}\}$ using a Bidirectional GRU (BiGRU) encoder with 128 dimensions (64 in each direction). To add temporal context to the dialogue history utterances, we append special positional tokens to each utterance.
\begin{equation} \label{eq:mnmemory}
m_k = BiGRU_m(\bm{u_k}) \quad for \quad 0\leq k\leq t-1
\end{equation}
We also encode the current utterance with another BiGRU encoder with 128 dimensions (64 in each direction), into a context vector denoted by $c$, as in equation ~\ref{eq:mncontext}. This is conceptually depicted in Figure~\ref{fig:memcontext}
\begin{equation} \label{eq:mncontext}
c = BiGRU_c(\bm{u_t})
\end{equation}
Let $M$ be a matrix with the $i$th row given by $m_i$. We obtain the cosine similarity between each memory vector, $m_i$, and the context vector $c$. The softmax of this similarity is used as an attention distribution over the memory $M$, and an attention weighted sum of $M$ is used to produce the dialogue context vector $h_t$ (Equation ~\ref{eq:mnattention}). This is conceptually depicted in Figure ~\ref{fig:cosinemem}.
\begin{equation} \label{eq:mnattention}
\begin{gathered}
a = softmax(Mc) \\
h_t = a^TM
\end{gathered}
\end{equation}

\subsubsection{Sequential Dialogue Encoder Network}
We enhance the memory network architecture described above by adding a session encoder~\cite{HRED2015} that temporally combines a joint representation of the current utterance encoding, $c$, (Eq. ~\ref{eq:mncontext}) and the memory vectors, $\{m_1, m_2 ... m_{t-1}\}$, (Eq. ~\ref{eq:mnmemory}). \\
We combine the context vector $c$ with each memory vector $m_k$, for $1 \leq k \leq n_k$, by concatenating and passing them through a feed forward layer (FF) to produce 128 dimensional context encodings, denoted by $\{g_1, g_2 ... g_{t-1}\}$ (Eq. ~\ref{eq:hcencontext}).
\begin{figure}
  \centering
	\includegraphics[width=0.8\textwidth]{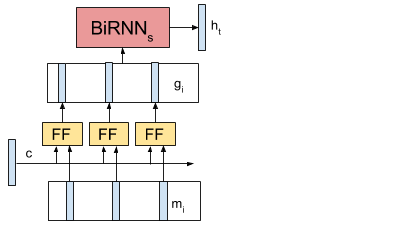}
	\caption{Architecture of the Sequential Dialogue Encoder Network. The feed-forward networks share weights across all memories.}
	\label{fig:hdenmem}
\vspace{-0.5cm}
\end{figure}
\begin{equation} \label{eq:hcencontext}
g_k = sigmoid(FF(m_k, c)) \quad for \quad 0\leq k\leq t-1
\end{equation}
These context encodings are fed as token level inputs into the session encoder, which is a 128 dimensional BiGRU layer. The final state of the session encoder represents the dialogue context encoding $h_t$ (Eq. ~\ref{eq:hcensession}).
\begin{equation} \label{eq:hcensession}
h_t = BiGRU_s(\{g_1, g_2, ... g_{t-1}\})
\end{equation}
The architecture is depicted in Figure ~\ref{fig:hdenmem}.
\subsection{Tagger Architecture}
\begin{figure*}
  \centering
	\includegraphics[width=0.8\textwidth]{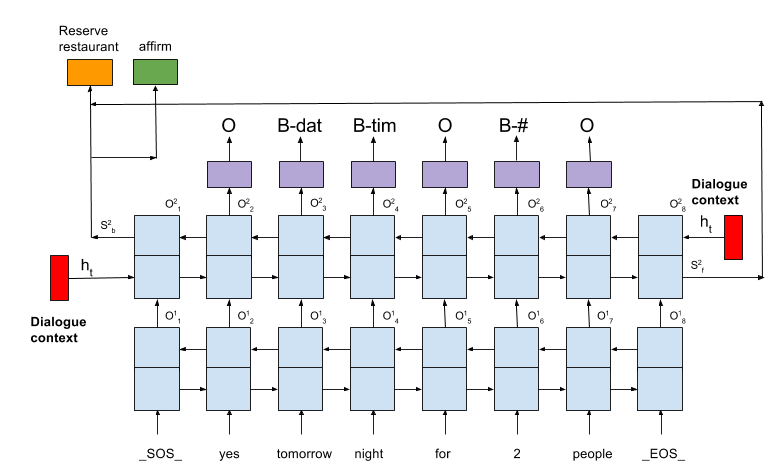}
	\caption{Architecture of the stacked BiRNN tagger. The dialogue context obtained from the context encoder is fed into the initial states of the second RNN layer.}
	\label{fig:multi}
\vspace{-0.5cm}
\end{figure*}
For all our experiments we use a stacked BiRNN tagger to jointly model domain classification, intent classification and slot-filling, similar to the approach described in ~\cite{dilekIS16}. We feed learned 256 dimensional embeddings corresponding to the current utterance tokens into the tagger. \\
The first RNN layer uses GRU cells with 256 dimensions (128 in each direction) as in equation ~\ref{eq:RNN1}. The token embeddings are fed into the token level inputs of the first RNN layer to produce the token level outputs $\bm{o^1} = \{o^1_1, o^1_2 ... o^1_{n_t}\}$.
\begin{equation} \label{eq:RNN1}
\bm{o^1}  = BiGRU_1(\bm{u_t})
\end{equation} 
The second layer uses Long Short Term Memory (LSTM) ~\cite{hochreiter1997long} cells with 256 dimensions (128 in both dimensions). We use a LSTM based second layer since that improved slot-filling performance on the validation set for all architectures. We apply dropout to the outputs of both layers. The initial states of both forward and backward LSTMs of the second tagger layer are initialized with the dialogue encoding $h_t$ as in equation ~\ref{eq:RNN2}. The token level outputs of the first RNN layer, $\bm{o^1}$, are fed as input into the second RNN layer to produce token level outputs $\bm{o^2} = \{o^2_1, o^2_2 ... o^2_{n_t}\}$ and the final state $s^2$.
\begin{equation} \label{eq:RNN2}
\bm{o^2}, s^2  = BiLSTM_2(\bm{o^1} , h_t)
\end{equation}
The final state of the second layer, $s^2$, is used as input to classification layers for domain and intent classification.
\begin{equation} \label{eq:domain}
\begin{gathered}
p^{domain} = softmax(Us^2) \\
p^{intent} = sigmoid(Vs^2)
\end{gathered}
\end{equation}
The token level outputs of the second layer, $\bm{o^2}$, are used as input to a softmax layer that outputs the IOB slot labels. This results in a softmax layer with $2N + 1$ dimensions for a domain with $N$ slots.
\begin{equation} \label{eq:slot}
p^{slot}_i = softmax(So^2_i) \quad for \quad 0\leq i\leq n^t
\end{equation}
The architecture is depicted in Figure ~\ref{fig:multi}.\\

\section{Dataset}

\begin{table*}[t]
\centering
\vspace{1cm}
\begin{tabular}{p{3cm}|p{10cm}}
\hline
\bf{Domain} & \bf{Attributes}\\
\hline
movies & date, movie, num\_tickets, theatre\_name, time \\ \hline
find-restaurants & category, location, meal, price\_range, rating, restaurant\_name\\ \hline
reserve-restaurant & date, num\_people, restaurant\_name, time \\ \hline
\end{tabular}
\caption{List of attributes supported for each domain.}
\label{tab:slotsem}
\end{table*}
\begin{table*}[t]
\centering
\begin{tabular}{p{4.5cm} | p{4.5cm}|p{4.5cm}}
\hline
\bf{Dialogue $x$} & \bf{Dialogue $y$} & \bf{Dialogue $d_r$}\\ \hline
U: Get me 5 tickets to see Inferno. & & U: Get me 5 tickets to see Inferno. \\ \hline
S: Sure, when is this booking for ? & & S: Sure, when is this booking for ? \\ \hline
U: Around 5 pm tomorrow night. & & U: Around 5 pm tomorrow night. \\ \hline
S: Do you have a theatre in mind? & & S: Do you have a theatre in mind? \\ \hline
U: AMC newpark 12. & U: Find italian restaurants in Mountain View & U: Find italian restaurants in Mountain View \\ \hline
S: Does 4:45 pm work for you ? & S: What price range are you looking for ? &  S: What price range are you looking for ? \\ \hline
U: Yes. & U: cheap & U: cheap \\ \hline
S: Your booking is complete. & S: Ristorante Giovanni is a nice Italian restaurant in Mountain View. & S: Ristorante Giovanni is a nice Italian restaurant in Mountain View. \\ \hline
& U: That works. thanks. & U: That works. thanks. \\ \hline
\end{tabular}
\caption{A sample dialogue obtained from recombining a dialogue from the movies and find-restaurant datasets.}
\label{tab:recombexample}
\end{table*}
We crowd sourced multi-turn dialogue sessions for 3 tasks: buying movie tickets, searching for a restaurant and reserving tables at a restaurant. Our data collection process comprises of two steps: (i) Generating user-agent interactions comprising of dialog acts and slots based on the interplay of a simulated user and a rule based dialogue policy. (ii) Using a crowd sourcing platform to elicit natural language utterances that align with the semantics of the generated interactions. \\
The goal of the spoken language understanding module of our dialogue system is to map each user utterance into frame based semantics that can be processed by the downstream components. Tables describing the intents and slots present in the dataset can be found in the appendix. \\
We use a stochastic agenda-based user simulator~\cite{schatzmann2007agenda,shah2016interactive} for interplay with our rule based system policy. The user goal is specified in terms of a tuple of slots, which denote the user constraints. Some constraints might be unspecified, in which case the user is indifferent to the value of those slots. At any given turn, the simulator samples a user dialogue act from a set of acceptable actions based on (i) the user goal and agenda that includes slots that still need to be specified, (ii) a randomly chosen user profile (co-operative/aggressive, verbose/succinct etc.) and (iii) the previous user and system actions. Based on the chosen user dialogue act, the rule based policy might make a backend call to inquire for restaurant or movie availability. Based on the user act and the backend response the system responds back with a dialogue act or a combination of dialogue acts, based on a hand designed rule based policy. These generated interactions were then translated to their natural language counterparts and sent out to crowd-workers for paraphrasing into natural language human-machine dialogues. \\
The simulator and policy were also extended to handle multiple goals spanning different domains. In this set-up, the user goal for the simulator would include multiple tasks and slot values could be conditioned on the previous task, for example, the simulator would ask for booking a table "after the movie", or search for a restaurant "near the theater".  The set of slots supported by the simulator is enumerated in Table ~\ref{tab:slotsem}.
We collected 1319 dialogues for restaurant reservation, 976 dialogues for finding restaurants and 1048 dialogues for buying movie tickets. All single domain datasets were used for training. The multi-domain simulator was used to collect 467 dialogues for training, 50 for validation and 273 for the test set. Since the natural language dialogues were paraphrased versions of known dialogue- act and slot combinations, they were automatically labeled. These labels were verified by an expert annotator, and turns with missing annotations were manually annotated by the expert.

\section{Dialogue Recombination}
As described in the previous section, we train our models on a large set of single domain dialogue datasets and a small set of multi-domain dialogues. These models are then evaluated on a test set composed of multi-domain dialogues, where the user attempts to fulfill multiple goals spanning several domains. This results in a distribution drift that might result in performance degradation. To counter this drift in the training-test data distributions we device a dialogue recombination scheme to generate multi-domain dialogues from single domain training datasets. \\
The key idea behind the recombination approach is the conditional independence of sub-dialogues aimed at performing distinct tasks ~\cite{grosz1986attention}. We exploit the presence of task intents, or intents that denote a switch in the primary task the user is trying to perform, since they are a strong indicator of a switch in the focus of the dialogue. We exploit the independence of the sub-dialogue following these intents from the previous dialogue context, to generate synthetic dialogues with multi-domain context. The recombination process is described as follows: \\
Let a dialogue $d$ be defined as a sequence of turns and corresponding semantic labels (domain, intent and slot annotations) $\{(t_{d1}, f_{d1}), (t_{d2}, f_{d2}), ... (t_{dn_d}, f_{dn_d}\}$. To obtain a re-combined dataset composed of dialogues from dataset $dataset_1$ and $dataset_2$, we repeat the following steps 10000 times, for each combination of $(dataset_1, dataset_2)$ from the three single domain datasets.
\begin{itemize}
\item Sample dialogues $x$ and $y$ from $dataset_1$ and $dataset_2$ respectively.
\item Find the first user utterance labeled with a task intent in $y$. Let this be turn $l$.
\item Randomly sample an insertion point in dialogue $x$. Let this be turn $k$.
\item The new recombined dialogue is $\{(t_{x1}, f_{x1}), ...  (t_{xk}, f_{xk}), (t_{yl},f_{yl}),$ $... (t_{yn_y}, f_{yn_y})\}$.
\end{itemize}
A sample dialogue generated using the above procedure is described in table ~\ref{tab:recombexample}. We drop the utterances from dialogue $x$ following the insertion point (turn $k$) in the recombined dialogue since these turns become ambiguous or confusing in the absence of preceding context. In a sense our approach is one of partial dialogue recombination.

\section{Experiments}
\begin{table*}
\centering
\begin{tabular}{p{2cm}|p{2cm}|p{2cm}|p{2.5cm}|p{3cm}}
\hline
\bf{Model} & \bf{Domain F1} & \bf{Intent F1} & \bf{Slot Token F1} & \bf{Frame Error Rate}\\ \hline
ED & 0.937 & 0.865 & 0.891 & 31.87\% \\ \hline
MN & 0.964 & 0.890 & 0.896 & 26.72\% \\ \hline
SDEN & 0.960 & 0.870 & 0.896 & 31.31\% \\ \hline
ED + DR & 0.936 & 0.885 & 0.911 & 30.72\% \\ \hline
MN + DR & 0.968 & \textbf{0.902} & 0.904 & 27.48\% \\ \hline
SDEN + DR & \textbf{0.975} & 0.898 & \textbf{0.926} & \textbf{25.85\%} \\ \hline
\end{tabular}
\caption{Test set performances for the encoder decoder (ED) model, Memory Network (MN) and the Sequential Dialogue Encoder Network (SDEN) with and without recombined data (DR).}
\label{tab:results}
\end{table*}

\begin{table*}[t]
\centering
\begin{tabular}{p{11cm} | p{2cm} | p{2cm}}
\hline
\textbf{utterance} &\textbf{MN+DR}  & \textbf{SDEN+DR}  \\ \hline
hi! & 0.00 &  0.13\\ \hline
hello ! i want to buy movie tickets for \textit{8} pm at cinelux plaza & 0.05  & \textbf{0.34}\\ \hline
which movie , how many , and what day ? & 0.13 &  \textbf{0.24} \\ \hline
\textit{Trolls} , \textit{6} tickets for today \\ \hline
\end{tabular}
\begin{tabular}{p{2cm}|p{3cm}|p{3cm}|p{3cm}|p{3cm}}
& \textbf{True} & \textbf{ED+DR} & \textbf{MN+DR} & \textbf{SDEN+DR}\\ \hline
\textbf{Domain} & buy-movie-tickets & movies & movies & movies\\ \hline
\textbf{Intent} & contextual & contextual& contextual& contextual \\ \hline
\textbf{date} & today & today & today & today\\ \hline
\textbf{num\_tickets} & \textit{6} & \textit{6} & \textit{6} & \textit{6} \\ \hline
\textbf{movie} &  \textit{Trolls} & \textit{Trolls} & - & \textit{Trolls}  \\
\end{tabular}
\caption{Dialogue from the test set with predictions from Encoder Decoder with recombined data (ED+DR), Memory Network with recombined data (MN+DR) and Sequential Dialogue Encoder Network with dialogue recombination (SDEN+DR).Tokens that have been italicized in the dialogue were out of vocabulary or replaced with special tokens. The columns to the right of the dialogue history detail the attention distributions. For SDEN+DR, we use the magnitude of the change in the session GRU state as a proxy for the attention distribution. Attention weights might not sum up to 1 if there is non-zero attention on history padding.}
\label{tab:successexample1}
\end{table*}
\begin{table*}[t]
\centering
\begin{tabular}{p{11cm} | p{2cm} | p{2cm}}
\hline \hline
\textbf{utterance}  &\textbf{MN+DR} & \textbf{SDEN+DR}\\ \hline
hello & 0.01 & 0.10\\ \hline
hello . i need to buy tickets at cinemark redwood downtown \textit{20} for xd at \textit{6} : \textit{00} pm & 0.00 & 0.06 \\ \hline
which movie do you want to see at what time and date . & 0.00 & 0.04\\ \hline
I didn't understand that. & 0.00 & 0.03\\ \hline
please tell which movie , the time and date of the movie & 0.01 & 0.02\\ \hline
the movie is queen of katwe today and the number of tickets is \textit{4} & 0.00 & 0.00\\ \hline
So 4 tickets for the \textit{6} : \textit{00} pm showing & 0.02 & 0.01\\ \hline
yes & 0.01 & 0.01\\ \hline
I bought you \textit{4} tickets for the \textit{6} : \textit{00} pm showing of queen of katwe at cinemark redwood downtown \textit{20} & 0.06& 0.04\\ \hline
thank you & 0.03 & 0.03 \\ \hline
i want a \textit{Brazilian} restaurant & \textbf{0.61} & \textbf{0.29}\\ \hline
which one of \textit{Fogo de Cho Brazilian} steakhouse , \textit{Espetus Churrascaria} san mateo or \textit{Fogo de Cho} would you prefer & 0.02 & \textbf{0.26}\\ \hline
\textit{Fogo de Cho Brazilian} steakhouse \\ \hline
\end{tabular}
\begin{tabular}{p{2cm}|p{3cm}|p{3cm}|p{3cm}|p{3cm}}
& \textbf{True} & \textbf{ED+DR}& \textbf{MN+DR} & \textbf{SDEN+DR}\\ \hline
\textbf{Domain} & find-restaurants & movies& find-restaurants & find-restaurants \\ \hline
\textbf{Intent} & affirm(restaurant) & - & - & - \\ \hline
\textbf{restaurant name}& \textit{Fogo de Cho Brazilian} steakhouse & - & - & \textit{Fogo de Cho Brazilian} steakhouse \\ \hline
\end{tabular}
\caption{Dialogue from the test set with predictions from Encoder Decoder with recombined data (ED+DR), Memory Network with recombined data (MN+DR) and Sequential Dialogue Encoder Network with dialogue recombination (SDEN+DR). Tokens that have been italicized in the dialogue were out of vocabulary or replaced with special tokens. The columns to the right of the dialogue history detail the attention distributions. For SDEN+DR, we use the magnitude of the change in the session GRU state as a proxy for the attention distribution. Attention weights might not sum up to 1 if there is non-zero attention on history padding.}
\label{tab:successexample2}
\end{table*}
We compare the domain classification, intent classification and slot-filling performances, and the overall frame error rates of the encoder-decoder, memory network and sequential dialogue encoder network on the dataset described above. The frame error rate of a SLU system is the percentage of utterances where it makes a wrong prediction i.e. any of domain, intent or slot is predicted incorrectly. \\
We trained all 3 models with RMSProp for 100000 training steps with a batch size of 100. We started with a learning rate of 0.0003 which was decayed by a factor of 0.95 every 3000 steps. Gradient norms were clipped if they exceed a magnitude of 2.5. All model and optimization hyper-parameters were chosen based on a grid search, to minimize validation set frame error rates.\\
We restrict the model vocabularies to contain only tokens occurring more than 10 times in the training set, to prevent over-fitting to training set entities. Digits were replaced with a special "\#" token to allow better generalization to unseen numbers. The dialogue history was padded to 40 utterances for batch processing. We report results with and without the recombined dataset in Table~\ref{tab:results}. 
\section{Results}
The encoder decoder model trained on just the previous turn context performs worst on almost all metrics, irrespective of the presence of recombined data. This can be explained by worse performance on in-dialogue utterances, where just the previous turn context isn't sufficient to accurately identify the domain, and in several cases, the intents and slots of the utterance.\\
The memory network is the best performing model in the absence of recombined data, indicating that the model is able to encode additional context effectively to improve performance on all tasks, even when only a small amount of multi-domain data is available. \\
The Sequential dialogue encoder network performs slightly worse than the memory network in the absence of recombined data. This could be explained by the model over-fitting to the single domain context seen during training and failure to utilize context effectively in a multi-domain setting. In the presence of recombined dialogues it outperforms all other implementations. \\
Apart from increasing the noise in the dialogue context, adding recombined dialogues to the training set increases the average turn length of the training data, bringing it closer to that of the test dialogues. Our augmentation approach is, in spirit, an extension of the data recombination described in ~\cite{jia2016data} to conversations. We hypothesize that the presence of synthetic context has a regularization-like effect on the models. Similar effects were observed by~\cite{jia2016data}, where training with longer, synthetically-augmented utterances resulted in improved semantic parsing performance on a simpler test set. This is also supported by the observation that performance improvements obtained by addition of recombined data increase as the complexity of the model increases.

\section{Discussion and Conclusions}
Table ~\ref{tab:successexample1} demonstrates an example dialogue from the test set, along with the gold and model annotations from all 3 models. We observe that Encoder Decoder (ED) and Sequential Dialogue Encoder Network (SDEN) are able to successfully identify the domain, intent and slots, while the Memory Network (MN) fails to identify the movie name. Looking at the attention distributions, we notice that the MN attention is very diffused, whereas SDEN is focusing on the most recent last 2 utterances, which directly identify the domain and the presence of the \textit{movie} slot in the final user utterance. ED is also able to identify the presence of a \textit{movie} in the final user utterance from the previous utterance context.\\
Table ~\ref{tab:successexample2} displays another example where the SDEN model outperforms both MN and ED. Constrained to just the previous utterance ED is unable to correctly identify the domain of the user utterance. The MN model correctly identifies the domain, using its strong focus on the task-intent bearing utterance, but it is unable to identify the presence of a restaurant in the user utterance. This highlights its failure to combine context from multiple history utterances. On the other hand, as indicated by its attention distribution on the final two utterances, SDEN is able to successfully combine context from the dialogue to correctly identify the domain and the restaurant name from the user utterance, despite the presence of several out-of-vocabulary tokens. \\
The above two examples hint that SDEN performs better in scenarios where multiple history utterances encode complementary information that could be useful to interpret user utterances. This is usually the case in more natural goal oriented dialogues, where several tasks and sub tasks go in and out of the focus of the conversation ~\cite{grosz1979focusing}. \\
On the other hand, we also observed that SDEN performs significantly worse in the absence of recombined data. Due to its complex architecture and a much larger set of parameters SDEN is prone to over-fitting in low data scenarios. \\
In this paper, we collect a multi-domain dataset of goal oriented human-machine conversations and analyze and compare the SLU performance of multiple neural network based model architectures that can encode varying amounts of context. Our experiments suggest that encoding more context from the dialogue, and enabling the model to combine contextual information in a sequential order results in a reduction in overall frame error rate. We also introduce a data augmentation scheme to generate longer dialogues with richer context, and empirically demonstrate that it results in performance improvement for multiple model architectures.

\section{Acknowledgements}
We would like to thank Pararth Shah, Abhinav Rastogi, Anna Khasin and Georgi Nikolov for their help with the user-machine conversation data collection and labeling. We would also like to thank the anonymous reviewers for their insightful comments.

\bibliography{dialogue_memory_paper.v2}{}
\bibliographystyle{acl_natbib}
\begin{table*}[p]
\centering
\caption{\textbf{Supported Intents: }List of intents and dialogue acts supported by the user simulator, with descriptions and representative examples. Acts parametrized with \textbf{slot} can be instantiated for any attribute supported within the domain.}
\begin{tabular}{p{3cm}|p{5cm}|p{5cm}}
\hline
\bf{Intent} & \bf{Intent descriptions} & \bf{Sample utterance}\\
\hline
affirm & generic affirmation & U: sounds good. \\
\hline
cant\_understand & expressing failure to understand system utterance & U: What do you mean ? \\
\hline
deny & generic negation & U: That doesn't work. \\
\hline
good\_bye & expressing end of dialogue & U: bye \\
\hline
thank\_you & expressing gratitude & U: thanks a lot! \\
\hline
greeting & greeting & U: Hi \\
\hline
request\_alts & request alternatives to a system offer & S: Doppio Zero is a nice italian restaurant near you. \newline U: Are there any other options available ? \\ \hline
affirm(\bf{slot}) & affirming values corresponding to a particular attribute & U: 5 pm sounds good to me.\\
\hline
deny(\bf{slot}) & negating a particular attribute. & U: None of those times would work for me. \\
\hline
dont\_care(\bf{slot}) & expressing that any value is acceptable for a given attribute & U: Any time should be ok. \\
\hline
movies & explicit intent to buy movie tickets & U: Get me 3 tickets to Inferno \\ \hline
reserve-restaurants & explicit intent to reserve a table at a restaurant & U: make a reservation at Max Brenner's \\ \hline
find-restaurants & explicit intent to search for restaurants & U: find cheap italian restaurants near me \\ 
\hline
contextual & implicit intent continuing from context, also used in place of inform & S: What time works for you ? \newline U: 5 pm tomorrow.\\ \hline
unknown\_intent & intents not supported by the dialogue system & U: What's the weather like in San Francisco ? \\
\hline
\end{tabular}
\label{tab:framesem}
\end{table*}
\begin{table*}[p]
\centering
\caption{\textbf{Sample dialogue: }Sample dialogue generated using a crowd working platform. The LHS consists of the instructions shown to the crowd workers based on the dialog act interactions between the user simulator and the rule based policy. The RHS describes the natural language dialog generated by the crowd workers.}
\begin{tabular}{p{7cm}|p{8cm}}
\hline
\bf{Instructions} & \bf{Natural language paraphrase}\\\hline
S: greetings & hi \\\hline
U:buy movie tickets AND date is DontCare & buy some movie tickets for any date \\\hline
S: provide preference for: theatre name & Which theatre would you like to make the reservation at ? \\\hline
U: theatre name is GB & the GB theatre \\\hline
S: provide preference for: time & When is the reservation for? \\\hline
U: number of tickets is 6 and time is DontCare & get 6 tickets for any time\\\hline
S: provide preference for: movie & Which movie would you like to go for ? \\\hline
U: movie is Power Rangers & The power rangers movie \\\hline
S: found the following: date is March 25 and time is 8:00 am & I found 6 tickets available for 8 am on March 25. \\\hline
U: disagree & That doesn't work for me \\\hline
S: found the following: date is March 25 and time is 6:00 pm & Does 6 pm on March 25 work for you ? \\\hline
U: accept & Sounds good, make the booking \\\hline
S: purchase confirmed with given details theatre name is GB and date is March 25 and time is 6:00 pm and number of tickets is 6 and movie is Power Rangers & Ok, I've reserved 6 tickets for the 6 pm showing of Power Rangers at the GB theatre on March 25. \\\hline
U: thank you & Thanks. \\\hline
\end{tabular}
\label{tab:sampledialog}
\end{table*}
\end{document}